\documentclass[10pt,twocolumn,letterpaper]{article}

\usepackage{cvpr}              
\usepackage{ulem} 
\definecolor{cvprblue}{rgb}{0.21,0.49,0.74}
\usepackage[pagebackref,breaklinks,colorlinks,allcolors=cvprblue]{hyperref}
\usepackage{utfsym, stix}
\usepackage{multirow}
\usepackage{booktabs}
\usepackage{makecell}
\usepackage{textcomp}
\usepackage{algorithm}
\usepackage{algorithmic}
\usepackage{ulem}
\def\paperID{11071} 
\def\confName{CVPR}
\def\confYear{2025}

\title{DnLUT: Ultra-Efficient Color Image Denoising via Channel-Aware Lookup Tables}

\author{Sidi Yang\\
\and
Second Author\\
}

\author{Sidi Yang$^1$ \quad
Binxiao Huang$^{1\dag}$ \quad
\text{Yulun Zhang$^2$} \quad
\text{Dahai Yu$^3$} \quad
\text{Yujiu Yang$^{4\dag}$} \quad
\text{Ngai Wong$^1$} \\ 
$^1$~The University of Hong Kong \quad
$^2$~Shanghai Jiaotong University \\
$^3$~TCL Corporate Research \quad
$^4$~Tsinghua University \\
\small \textit{\{yangsidi99, huanghx7\}@connect.hku.hk,~yulun100@gmail.com,~dahai.yu@tcl.com}\\
                \small \textit{yang.yujiu@sz.tsinghua.edu.cn,~nwong@eee.hku.hk}}

\begin{document}

\maketitle
\renewcommand{\thefootnote}{$\dag$} 
\footnotetext{~Corresponding authors.} 
\begin{abstract}
\label{abs}
    While deep neural networks have revolutionized image denoising capabilities, their deployment on edge devices remains challenging due to substantial computational and memory requirements. To this end, we present DnLUT, an ultra-efficient lookup table-based framework that achieves high-quality color image denoising with minimal resource consumption. Our key innovation lies in two complementary components: a Pairwise Channel Mixer (PCM) that effectively captures inter-channel correlations and spatial dependencies in parallel, and a novel L-shaped convolution design that maximizes receptive field coverage while minimizing storage overhead. By converting these components into optimized lookup tables post-training, DnLUT achieves remarkable efficiency - requiring only 500KB storage and 0.1\% energy consumption compared to its CNN contestant DnCNN, while delivering 20× faster inference. Extensive experiments demonstrate that DnLUT outperforms all existing LUT-based methods by over 1dB in PSNR, establishing a new state-of-the-art in resource-efficient color image denoising. The project is available at \url{https://github.com/Stephen0808/DnLUT}.
\end{abstract}
\section{Introduction}
\label{sec:intro}

Color images pervade our digital world - from social media and photography to scientific imaging and medical diagnostics. These images are inherently more vulnerable to noise contamination than their grayscale counterparts, not only due to their threefold data size during acquisition, storage, and transmission, but also because human perception exhibits heightened sensitivity to color distortions compared to luminance variations.

The advent of deep neural networks (DNNs) has dramatically advanced the field of color image denoising\cite{zhang2017beyond, liang2021swinir, zamir2022restormer, chen2021pre, zhang2023kbnet}. State-of-the-art models leverage sophisticated architectures with numerous convolutional layers or transformer blocks, achieving unprecedented denoising quality. However, these advances come at a steep cost - the computational complexity and memory requirements of such models render them impractical for edge devices. This limitation is particularly acute given that most edge devices lack specialized hardware accelerators such as graphics processing units (GPUs) or tensor processing units (TPUs), creating a significant gap between algorithmic capabilities and practical deployability.

\begin{figure}
    \centering
    \includegraphics[width=0.98\linewidth]{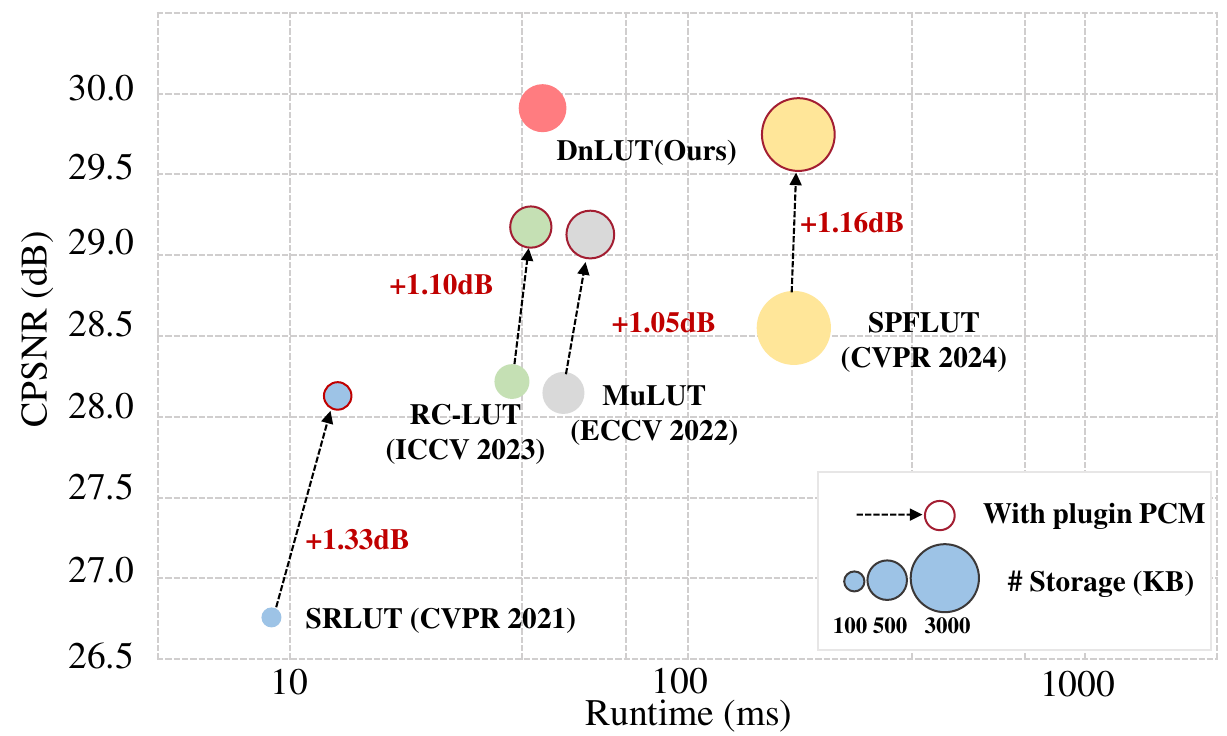}
    \caption{Model comparison in terms of color peak signal-to-noise ratio (CPSNR), runtime, and storage. The CPSNR and runtime are calculated on the CBSD68 dataset with Gaussian noise level $\sigma = 25$ using Qualcomm Snapdragon 8 Gen2. Our method outperforms state-of-the-art LUT-based methods with the highest CPSNR, at a low storage requirements and reduced runtime. Additionally, our PCM module serves as a versatile plug-in module that enhances existing methods' performance by over 1dB.}
    \label{fig:enter-label}
\end{figure}

Recent works have explored lookup tables (LUTs) as an elegant solution for efficient image enhancement. LUTs replace complex runtime computations with simple array indexing operations through direct addressing. This approach typically involves training a neural network model, then converting it into LUTs by exhaustively mapping input-output relationships. During inference, the system performs rapid table lookups instead of costly computations, effectively leveraging DNN capabilities while circumventing hardware constraints. However, LUT-based methods face a fundamental challenge: the storage requirement grows exponentially with the input dimensionality. This has led to the adoption of 4D LUTs (using four pixel values for indexing) as a practical compromise between storage and performance.

Current LUT-based methods predominantly employ $2\times2$ convolutional kernels processed independently for each channel\cite{jo2021practical, li2022mulut, liu2023reconstructed, li2024look}, creating spatial-wise LUTs for single-channel processing. While this approach works well for tasks like super-resolution, it proves inadequate for color image denoising where noise typically affects all RGB channels simultaneously. Channel correlation plays a pivotal role in modeling noise characteristics, a critical factor for achieving effective restoration. Alternative approaches either use $1\times1$ convolutions with three-channel depth for RGB information (creating 3D channel-wise LUTs)\cite{li2024toward, yang2024taming} or attempt to use larger $H\times W\times3$ kernels. However, these solutions either disrupt noise distribution patterns or lead to prohibitive storage requirements.

To address these challenges, we propose DnLUT, a novel lookup table-based framework specifically designed for efficient color image denoising. Our approach centers on two key innovations:

First, we introduce the Pairwise Channel Mixer (PCM), which simultaneously processes spatial and channel information. PCM strategically combines RGB channels into three pairs (RG, GB, BR), processing them through parallel branches using a $1\times2$ convolution kernel with depth 2. This design efficiently captures channel correlations while maintaining manageable storage requirements. The PCM module can also serve as a plug-in module for existing LUT-based methods, consistently improving their performance by approximately 1dB.

Second, we develop an innovative L-shaped convolution kernel that addresses the limitations of conventional rotation-based approaches. After taking the channel correlation into account, as discussed in \cite{li2022mulut, li2024look}, there is still a need to enlarge the spatial receptive field. While previous methods like SR-LUT\cite{jo2021practical} use rotation ensemble training to expand receptive fields of spatial-wise LUT, they often introduce redundant pixel usage and significant storage overhead. Our L-shaped kernel design eliminates pixel overlap during rotations, enabling conversion to more efficient 3D LUTs while maintaining the same effective receptive field size as 4D LUTs.

Extensive experimental validation demonstrates the superiority of DnLUT over existing approaches:
\begin{itemize}
\item We achieve up to 1.3dB CPSNR improvement over SPFLUT\cite{li2024look} on standard Gaussian denoising benchmarks. For real-world denoising tasks, we outperform CBM3D\cite{dabov2007color} by nearly 5dB.
\item Our method maintains exceptional efficiency, requiring only $0.1\%$ of DnCNN\cite{zhang2017beyond}'s energy consumption and 500KB storage.
\item The PCM module serves as a versatile performance enhancer, bringing over 1dB CPSNR improvement to existing LUT-based methods.
\item Our L-shaped convolution design reduces storage requirements by 17× while preserving receptive field coverage.
\end{itemize}

These results establish DnLUT as a significant advance in practical, resource-efficient color image denoising, bridging the gap between algorithmic capability and deployability on edge devices.
\section{Related works}
\label{sec:related}

\subsection{Color Image denoising}
A fundamental characteristic of color images is the strong correlation among RGB channels\cite{dai2013multichannel}. Research has consistently shown that processing these channels independently yields suboptimal results compared to joint processing approaches\cite{mairal2007sparse}. This insight has driven numerous innovations in color image denoising.

A landmark development was CBM3D\cite{dabov2007color}, which operates in luminance-chrominance space (\textit{i.e}\onedot, $YC_bC_r$). The method performs BM3D\cite{dabov2007image} grouping once on the luminance channel $Y$ and leverages this grouping for collaborative filtering of chrominance channels $U$ and $V$. Building on this concept, \cite{lefkimmiatis2017non} introduced a radial basis function mixture for more sophisticated handling of channel correlations. Another notable approach, MC-WNNM\cite{xu2017multi}, extended WNNM\cite{gu2014weighted} by incorporating a weight matrix to adaptively balance the data fidelity across channels based on their noise characteristics.

The deep learning era, initiated by DnCNN\cite{zhang2017beyond}, transformed the field of color image denoising. Subsequent approaches\cite{zamir2021multi, zamir2022restormer, liang2021swinir, chen2021pre, zhang2023kbnet} have increasingly sophisticated architectures that process all RGB channels simultaneously through complex networks of convolution layers or transformer blocks. While these models achieve remarkable denoising quality, their substantial computational and memory requirements present significant deployment challenges, particularly for resource-constrained edge devices.

\subsection{Replacing CNN with LUT}
The challenge of deploying deep neural networks on edge devices has sparked interest in lookup table (LUT) based alternatives. Inspired by the prevalence of color LUTs in image signal processors, \cite{zeng2020learning} pioneered an approach using learnable basis LUTs combined with a weight prediction network for image enhancement. \cite{yang2024taming} further developed this concept, proposing a more efficient channel-wise LUT system. However, these early channel-wise LUTs, processing single pixels per channel, proved inadequate for the complex task of denoising.

A significant breakthrough came with \cite{jo2021practical}, which introduced a method to convert CNNs with limited receptive fields into spatial-wise LUTs for super-resolution. This approach trains networks on restricted pixel neighborhoods (\textit{i.e}\onedot., $2\times2$) and converts them to LUTs by exhaustively mapping input-output relationships. During inference, the system efficiently retrieves pre-computed outputs through coordinate indexing. Follow-up works\cite{ma2022learning, li2022mulut, liu2023reconstructed} enhanced this framework using serial LUTs to expand receptive fields, achieving substantial performance improvements.

To manage the exponential growth of LUT size with input dimensionality, these methods typically process each input channel independently. While this approach proves effective for super-resolution, where spatial information dominates, it falls short for color image denoising, which requires preserving both spatial and channel information to model noise distribution effectively. The limitations of existing channel-wise and spatial-wise LUT approaches highlight the need for a more sophisticated solution that can handle the unique challenges of color image denoising while maintaining computational efficiency.
\section{Method}
\subsection{Preliminary}
The foundation of LUT-based image restoration was established by SR-LUT\cite{jo2021practical}, which introduces a constrained receptive field approach for network training (\textit{i.e}\onedot, $2\times2$). Post-training, the network's output values are systematically cached into a LUT by exhaustively traversing all possible input pixel value combinations. During inference, input patches of size $2\times2$ are converted to indices of a 4D LUT to retrieve corresponding output values. To manage storage requirements, the 4D index space is uniformly subsampled. Additionally, a rotation ensemble strategy rotates each $2\times2$ input patch four times, effectively expanding the receptive field to $3\times3$. Recent works have focused on increasing the effective receptive field through various techniques, including multiple LUTs\cite{li2022mulut, ma2022learning, li2024look}, shift aggregations\cite{ma2022learning}, and diverse kernel patterns\cite{li2022mulut, liu2023reconstructed, huang2023hundred}.

Our work advances this foundation by specifically addressing the challenges of LUT-based color image denoising, a critical application for edge devices.

A notable limitation of existing LUT-based approaches is their treatment of channel depth, typically set to 1. For color images, this means each RGB channel independently accesses cached output values without inter-channel interaction. This design overlooks crucial RGB channel correlations and spatial-channel-wise degradation patterns inherent in color image denoising. While a straightforward solution might be to employ vanilla convolution kernel patterns, the exponential growth of LUT size with index dimension makes this impractical. For instance, a kernel of size $2\times2$ with depth 3 would require approximately 582 TB of storage, even with subsampling compression. This constraint presents a fundamental dilemma: \textbf{when aiming to capture more spatial information, we have to sacrifice the richness of channel information.} Moreover, while 4D spatial-wise LUTs show promise, they typically require multiple LUTs (analogous to stacked convolution layers), creating significant storage overhead, particularly problematic for L1 cache access.

\begin{table}[th]
    \centering
    \caption{LUT size versus receptive field (RF) and channel numbers. * means subsampled LUT size}
    \begin{tabular}
    {p{10mm}<{\centering} p{10mm}<{\centering} p{15mm}<{\centering} p{25mm}<{\centering}}
        \toprule[1.1pt]
        RF& Depth &LUT Dim.& LUT Size*\\
        \hline
        $1\times1$ & 1 & 1D &17 B\\
        $1\times1$& 3& 3D &4.9 KB\\
        $2\times2$ & 1 & 4D &83.5 KB\\
        $1\times2$ & 3 & 6D &24.1 MB\\
        $2\times2$ & 3 & 12D & 582.6 TB\\
        $k\times k$ & c & $k \times k \times c$D & $(2^{4}+1)^{k \times k \times C}$ B\\
        \bottomrule[1.1pt]
    \end{tabular}
    \vspace{-1.0em}
    \label{tab:tradeoff}
\end{table}

\begin{figure*}
    \centering
    \includegraphics[width=0.98\linewidth]{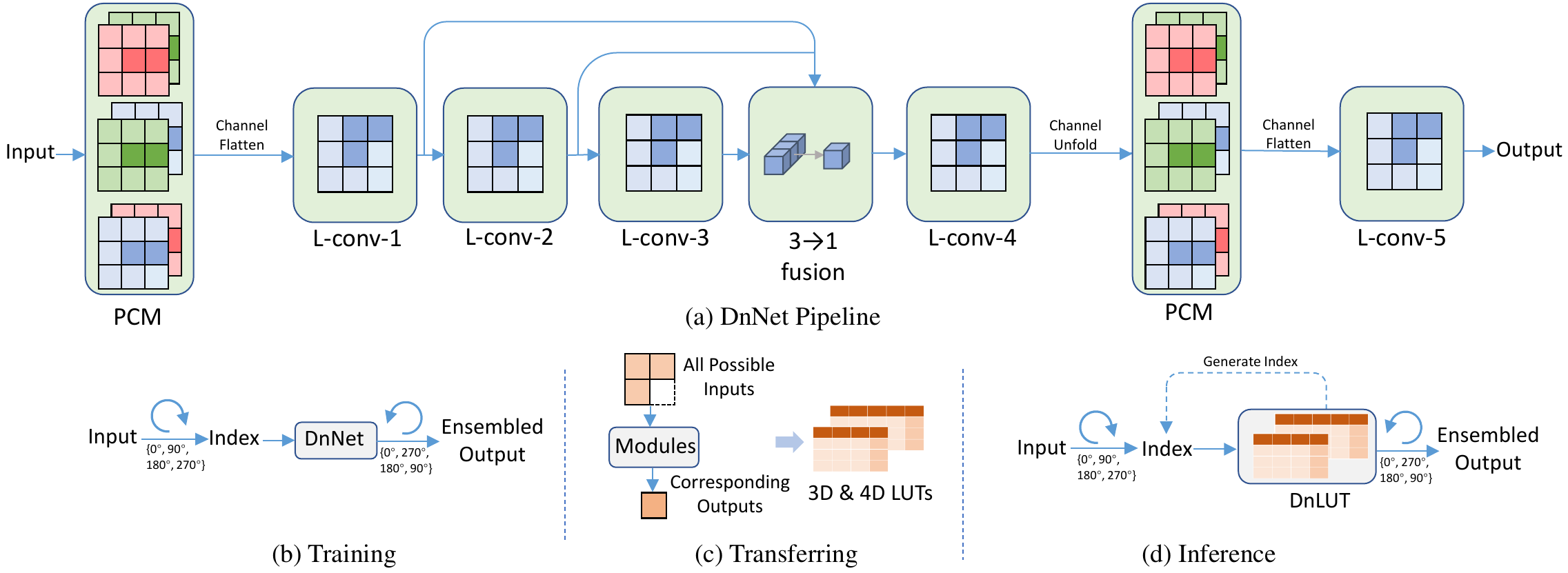}
    \caption{System architecture of DnLUT: (a) The DnNet pipeline integrates pairwise channel mixers and L-shaped convolutions, with multi-scale fusion enhancing receptive field coverage. Channel dimensions are flattened for parallel processing in L-shaped operations, then unfolded for PCM input. (b) Input pixels undergo four rotations (0\textdegree, 90\textdegree, 180\textdegree, 270\textdegree) during processing, with outputs averaged for enhanced results. (c) Post-training, all possible input combinations are processed through DnNet modules, with outputs cached in optimized 3D or 4D LUTs. (d) During inference, input pixels are efficiently processed through multiple LUTs, with each LUT's outputs informing subsequent LUT indices, culminating in final denoised pixel values.}
    \label{fig:pipeline}
\end{figure*}

\subsection{Overview}
To address these fundamental challenges, we introduce two complementary innovations: 1) a pairwise channel mixer that leverages 4D LUT efficiency while modeling spatial-channel correlations; and 2) a novel L-shaped kernel featuring inherent rotation non-overlapping properties, enabling conversion to 3D LUT while maintaining 4D LUT-equivalent receptive field coverage.

We first build the DnNet for training, as shown in Fig.~\ref{fig:pipeline}. Our proposed framework enhances traditional spatial-wise LUT capabilities through these two key modules: the pairwise channel mixer (PCM) establishes robust channel-spatial correlations in three RGB channels while the L-shaped kernel expands the effective receptive field with minimal computational overhead independently in each channel. The architecture combines PCM with L-shaped convolution in two stages.  The first stage, inspired by \cite{li2024look}, generates multi-channel features from LUT groups, which are then concatenated in the fusion module. The second stage integrates both proposed modules. Importantly, all components are designed for efficient conversion to 3D or 4D LUTs during inference, collectively referred to as DnLUT.

\subsection{Pairwise Channel Mixer}
\begin{figure}
    \centering
    \includegraphics[width=0.98\linewidth]{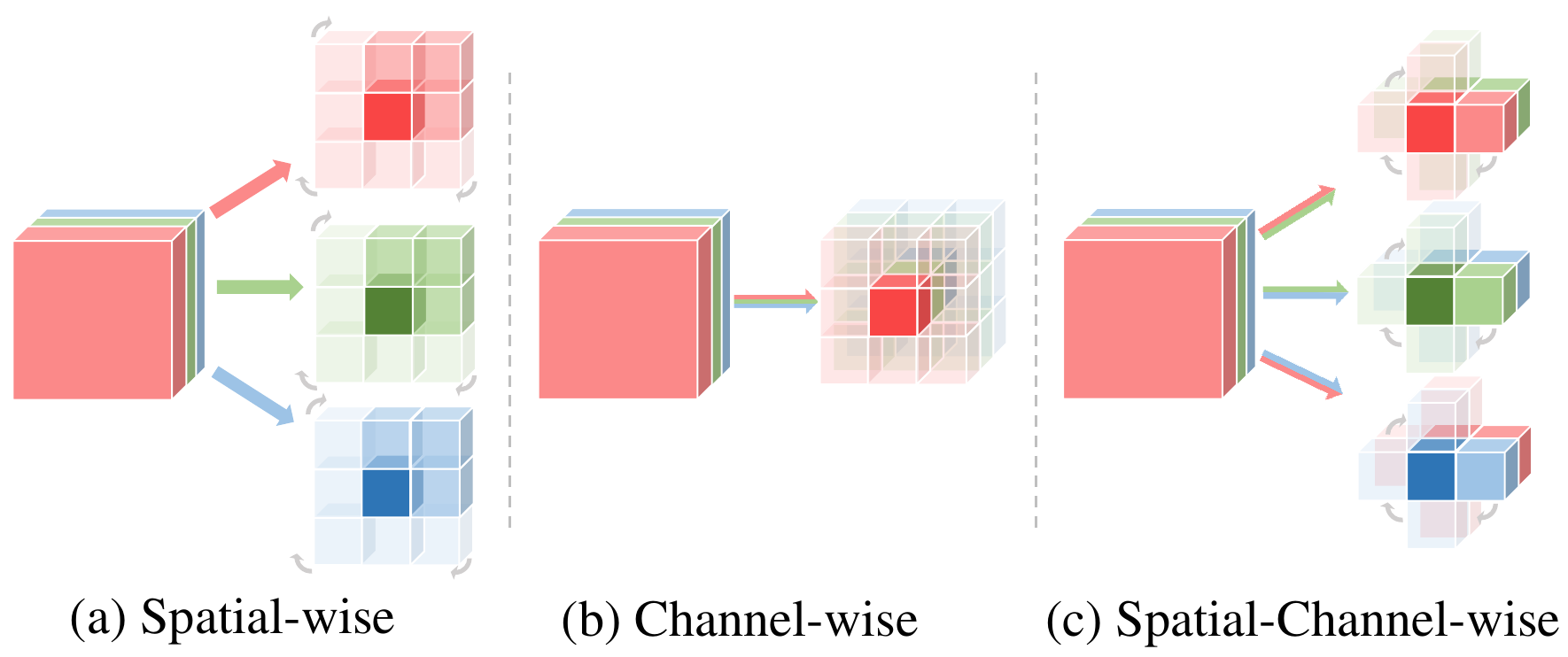}
    \caption{Taxonomy of kernel patterns for LUT-based methods. Dark cubes indicate rotation points, while medium-dark regions show involved pixel positions during one rotation.}
    \label{fig:diff_wise}
    \vspace{-0.8em}
\end{figure}

Designing effective kernel patterns for LUT-based methods presents significant challenges. Fig.~\ref{fig:diff_wise} illustrates various kernel pattern categories. Traditional approaches have focused on single-dimension kernels due to LUT index dimension constraints, leading to two major limitations in color image denoising. Spatial-wise kernels, while capturing spatial information, neglect channel correlations, resulting in color distortion and suboptimal noise modeling. Conversely, channel-wise kernels establish strong RGB channel connections but ignore spatial relationships, leading to persistent artifacts. While combining these kernels in a cascade network might seem promising, this approach fails to process channel and spatial information simultaneously, resulting in suboptimal noise distribution modeling.

Our PCM addresses these limitations through a novel architecture for color image denoising. The module first reorganizes RGB channels into three pairwise combinations: RG, GB, and BR. These pairs are processed through parallel branches using a kernel with $1\times2$ spatial dimensions and depth 2 in the initial layer, followed by cascaded $1\times1$ convolution layers. Each convolution operation processes four pixel values to produce one channel output, enabling efficient conversion to 4D LUTs post-training. The cached LUTs (\textit{i.e}\onedot, $LUT_{RG}, LUT_{GB}, LUT_{GR}$) operate in parallel, with their predictions concatenated. For an input anchor $(I_{R}, I_{G}, I_{B})$, the output values $(V_{R}, V_{G}, V_{B})$ are computed as:
\begin{equation}
\begin{aligned}
    (V_{R}, V_{G}, V_{B}) &= \text{Cat}(LUT_{RG}[I_{0,0,R}][I_{0,1,R}][I_{0,0,G}][I_{0,1,G}], \\
                           &\quad \quad \quad LUT_{GB}[I_{0,0,G}][I_{0,1,G}][I_{0,0,B}][I_{0,1,B}], \\
                           &\quad \quad \quad LUT_{BR}[I_{0,0,B}][I_{0,1,B}][I_{0,0,R}][I_{0,1,R}]),
\end{aligned}%
\end{equation}%
where $H, W, C$ of each $I_{H,W,C}$ denotes height, width, and channel index. PCM effectively balances spatial and channel-wise processing, capturing inter-channel correlations crucial for color fidelity while preserving spatial relationships through the $1\times2$ kernel design and refining features via cascaded $1\times1$ convolutions.

\subsection{Rotation Non-overlapping Kernel}
\begin{figure}
    \centering
    \includegraphics[width=0.85\linewidth]{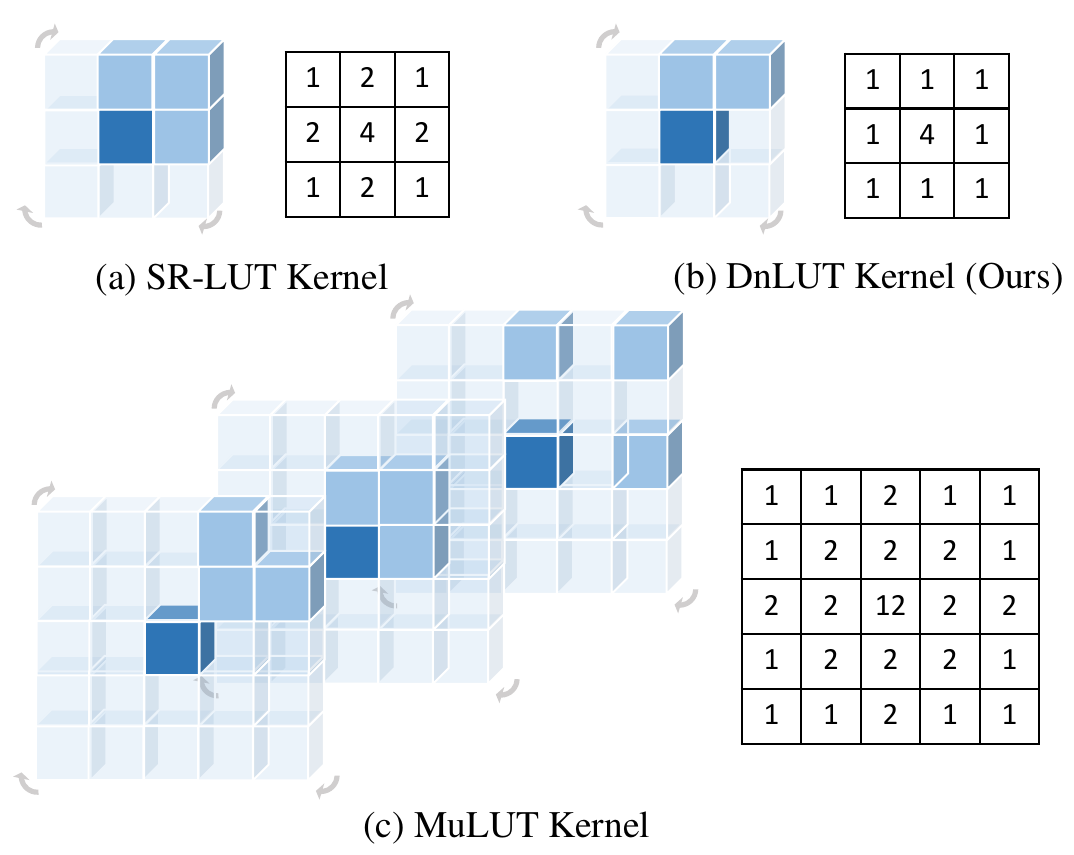}
    \caption{Comparison of spatial-wise kernel designs. Left patterns show kernel configurations, while right tables quantify lookup frequencies during output retrieval.}
    \label{fig:non-overlapping}
    \vspace{-1.0em}
\end{figure}

The rotation ensemble strategy is fundamental to LUT-based methods, using four rotations (0\textdegree, 90\textdegree, 180\textdegree, 270\textdegree) to enable asymmetric convolution kernels to cover larger symmetric areas. By combining outputs from all rotations, the approach enhances performance without increasing LUT index dimensionality. As shown in Fig.~\ref{fig:non-overlapping}, traditional methods like SR-LUT and MuLUT use this strategy to expand receptive fields to $3\times3$ and $5\times5$ respectively.

However, this conventional approach has two critical limitations. First, approximately half of the pixels within the receptive field are accessed multiple times during lookup operations, as illustrated in Fig.~\ref{fig:non-overlapping}'s tables. This redundancy indicates inefficient kernel pattern design. Second, stacking these kernels to create multiple LUTs for expanded receptive fields results in significant 4D LUT storage overhead.

To address these inefficiencies, we introduce an innovative L-shaped kernel design. During each rotation, the kernel processes two additional pixels (beyond the center pixel) without overlap, ensuring each surrounding pixel contributes exactly once to the output. This design maximizes pixel value utilization within the receptive field while enabling conversion to 3D LUTs instead of 4D LUTs, dramatically reducing storage requirements without sacrificing coverage.

\subsection{PCM Plug-in Module}
To demonstrate the versatility of our pairwise channel mixer, we developed PCM as a plug-in module compatible with existing LUT-based methods. The inference of plug-in algorithm is summarized in Alg.~\ref{alg:pcm}. This module seamlessly integrates with current architectures, enhancing their ability to capture channel-spatial correlations without requiring structural modifications. Compared to state-of-the-art methods like SPFLUT, the PCM plug-in adds only 12\% runtime and 8\% storage overhead while delivering over 1dB performance improvement. 

\begin{algorithm}
\caption{Inference of Pairwise Channel Mixer Module}
\begin{algorithmic}[1]
\REQUIRE Input image $I$ of size $H \times W \times 3$
\REQUIRE 4D Lookup Table $LUT$ of PCM
\REQUIRE Padding size $p$

\STATE Initialize output image $O$ of size $H \times W \times 3$

\FOR{$r = 0$ to $3$}
    \STATE Rotate image $I$ by $r \times 90$ degrees to get $I_{\text{rot}}$
    \STATE Pad $I_{\text{rot}}$ with $p$ pixels on each side to get $I'_{\text{rot}}$
    
    \FOR{$h = 0$ to $H-1$}
        \FOR{$w = 0$ to $W-1$}
            \FOR{$c = 0$ to $2$}
                \STATE $p_1 \gets I'_{\text{rot}}[h, w, c]$
                \STATE $p_2 \gets I'_{\text{rot}}[h, w+1, c]$
                \STATE $p_3 \gets I'_{\text{rot}}[h, w, (c+1) \mod 3]$
                \STATE $p_4 \gets I'_{\text{rot}}[h, w+1, (c+1) \mod 3]$
                
                \STATE $index \gets \text{Quantize}(p_1, p_2, p_3, p_4)$
                \STATE $value \gets LUT[index]$
                \STATE $O[h, w, c] \gets O[h, w, c] + \text{Interpolate}(value)$
            \ENDFOR
        \ENDFOR
    \ENDFOR
    
    \STATE Rotate $O$ back by $(4 - r) \times 90$ degrees to get $O_{\text{rot}}$
    \STATE $O \gets O + O_{\text{rot}}$
\ENDFOR

\STATE $O \gets O / 4$

\RETURN Output image $O$
\end{algorithmic}
\label{alg:pcm}
\end{algorithm}

\section{Experiments}

\begin{table*}[htp]
    \centering
    \caption{Quantitative comparison on Gaussian color image denoising benchmark. The table presents CPSNR(dB) values.}
    \resizebox{2\columnwidth}{!}{
    \setlength{\tabcolsep}{1.5mm}{
    \begin{tabular}{ccccccccccccccccc}
        \toprule[1.1pt]
        \multirow{2}{*}{Category} & \multirow{2}{*}{Method}  & \multicolumn{3}{c}{CBSD68} & \multicolumn{3}{c}{Kodak24}&   \multicolumn{3}{c}{Urban100} &\multicolumn{3}{c}{McMaster}\\
        \cmidrule(r){3-5}
        \cmidrule(r){6-8}
        \cmidrule(r){9-11}
        \cmidrule(r){12-14}
        & & $\sigma = 15$ & $\sigma = 25$ &$\sigma = 50$ &$\sigma = 15$ & $\sigma = 25$ &$\sigma = 50$ &$\sigma = 15$ & $\sigma = 25$ &$\sigma = 50$ &$\sigma = 15$ & $\sigma = 25$ &$\sigma = 50$ \\
        \hline
        \multirow{5}{*}{LUT-based}&SR-LUT\cite{jo2021practical}  &29.76 & 26.71& 22.41&30.35 & 27.16& 22.65& 29.38 & 26.04& 21.60&31.18 & 28.01&23.35\\
         &MuLUT\cite{li2022mulut}  &30.52 & 28.11& 24.85& 31.31& 29.02& 25.28 & 30.25& 27.67& 23.75& 32.28& 29.88& 26.36\\
         &RC-LUT\cite{liu2023reconstructed}  &30.68 & 28.12&25.04 & 31.57 & 29.07& 25.89&30.33 & 27.80 & 23.86& 32.51&29.89&26.50\\
         &SPF-LUT\cite{li2024look}  & 30.97 & 28.56 &25.33 & 31.86 & 29.58 &26.26 & 30.89 & 28.26& 24.22 &31.77 & 30.44 & 26.91\\
         &DnLUT(Ours) & 32.41 & 29.88 & 26.03 &  33.02 & 30.24 & 26.74 & 32.12 & 28.87 & 25.01 & 32.88 & 30.44& 27.12\\ 
\midrule
         \multirow{2}{*}{Classical}&CBM3D\cite{dabov2007color} &33.52 & 30.71 & 27.38 & 34.28 & 31.68 & 29.02 &33.93 & 31.36&27.93&34.06 & 31.66 & 28.51 \\
         &MC-WNNM\cite{xu2017multi} &31.98 & 29.32 & 26.98 & 33.23 & 30.89 & 28.67 & 30.23 & 29.23& 27.00& 31.23 & 30.20 & 27.55\\
\midrule
         \multirow{2}{*}{DNN}&DnCNN\cite{zhang2017beyond} &33.90 & 31.24 & 27.95 & 34.60 & 32.14 & 28.96& 32.98 & 30.81 & 27.59 & 33.45 & 31.52 & 28.62 \\
         &SwinIR\cite{liang2021swinir} & 34.42 & 31.78 & 28.56 & 35.34 & 32.89 & 29.79 & 35.61 & 33.20 & 30.22 & 35.13 & 32.90 & 29.82\\
        \bottomrule[1.1pt]
        \vspace{-1.0em}
        
    \end{tabular}}
    }
    \label{tab:benchmark}
\end{table*}

\begin{table*}[htp]
    \centering
    \caption{Quantitative comparison on real-world color image denoising. The table presents CPSNR(dB) and SSIM values. Methods on DnD dataset are validated on the online platform. We could only get the PSNR rather than CPSNR.}
    \setlength{\tabcolsep}{1.5mm}{
    \resizebox{2\columnwidth}{!}{
    \begin{tabular}{cccccccccccccccc}
        \toprule[1.1pt]
        \textbf{Datasets} & Method & SR-LUT\cite{jo2021practical} & MuLUT\cite{li2022mulut} & RC-LUT\cite{liu2023reconstructed} & SPF-LUT\cite{li2024look} & DnLUT(Ours) & CBM3D\cite{dabov2007color} & MC-WNNM\cite{xu2017multi} & DnCNN\cite{zhang2017beyond} \\
        \hline
        \multirow{2}{*}{SIDD} & CPSNR &29.38 & 33.24 & 33.88 & 34.91 & 35.44 & 30.14 & 29.45 & 36.45 \\
         & SSIM & 0.634 & 0.830 & 0.839 & 0.865 & 0.875 & 0.702 & 0.682 & 0.900\\
         \hline
         \multirow{2}{*}{DnD} & PSNR$^{*}$ & 33.69&35.11 & 35.43& 36.22& 36.67 & 33.12& 32.34& 37.11\\
         
         & SSIM &0.839 & 0.868& 0.875&0.911 & 0.922 & 0.823 & 0.817& 0.932\\
        \bottomrule[1.1pt]
        \vspace{-1.0em}
        
    \end{tabular}}
    }
    \label{tab:sidd}
\end{table*}

\begin{figure*}
    \centering
    \includegraphics[width=0.98\linewidth]{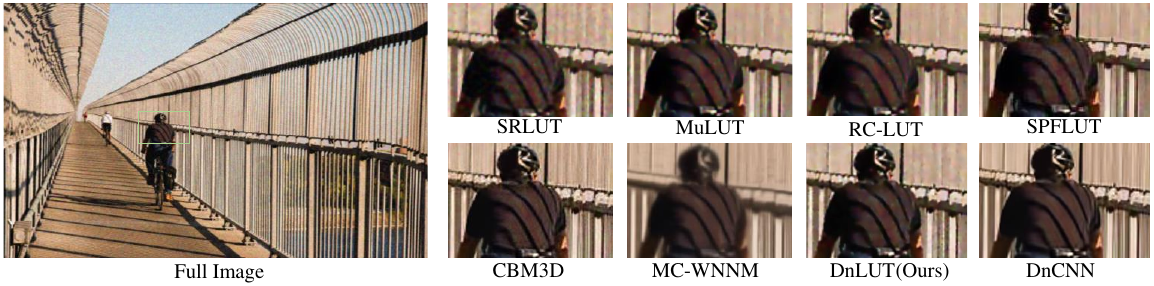}
    \caption{Qualitative comparison on synthetic datasets.}
    \label{fig:Gua}
\end{figure*}

\begin{figure*}
    \centering
    \includegraphics[width=0.98\linewidth]{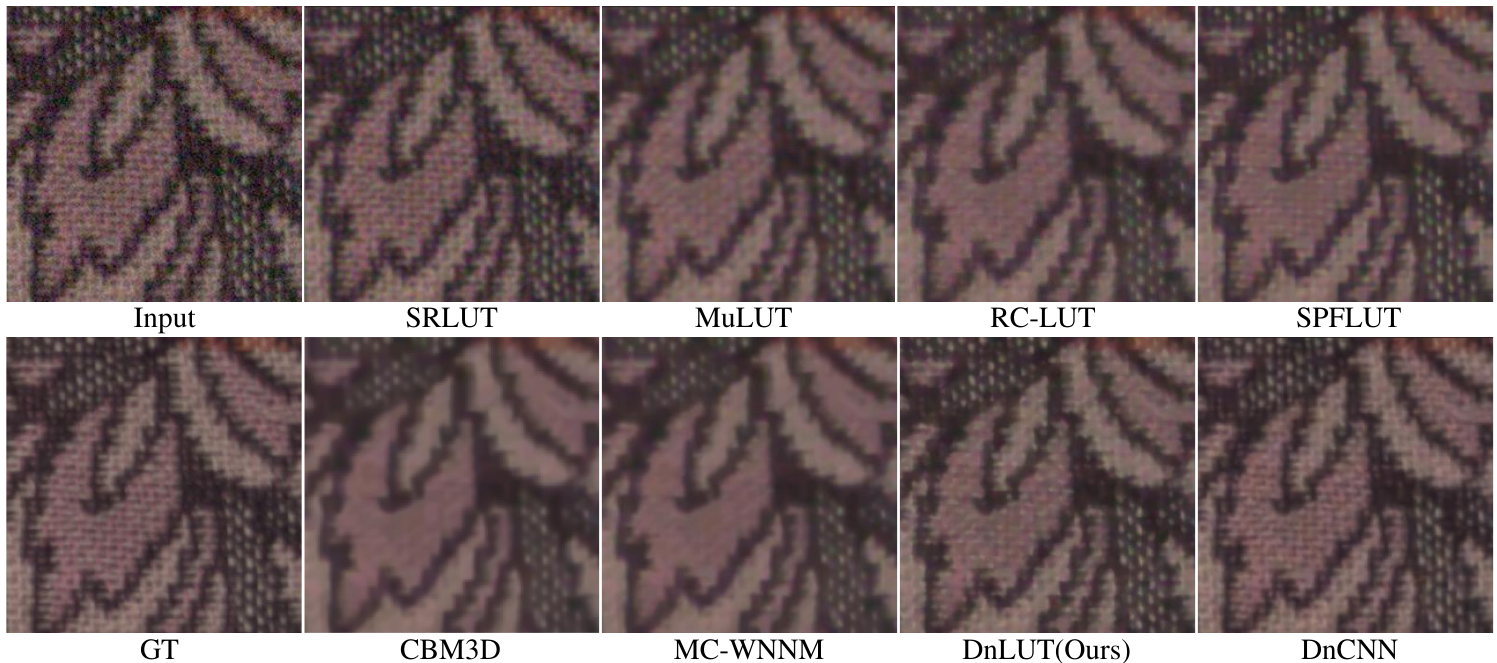}
    \caption{Qualitative comparison on real-world datasets.}
    \label{fig:sidd}
\end{figure*}

\begin{table*}[htp]
    \centering
    \caption{PCM can be easily incorporated to the existing LUT-based methods. It equips the model with the ability of capturing the channel correlations which  brings more than 1dB on the widely used benchmark. The noise level $\sigma$ of Gaussian noise is set to 25. The table presents CPSNR(dB) values.}
    \begin{tabular}{cccccc}
        \toprule[1.1pt]
        Method& CBSD68 & Kodak24 & Urban100 & McMaster & SIDD \\
        \hline
        SR-LUT\cite{jo2021practical} & 26.71 & 27.16 & 26.04 & 28.01 &29.38\\
        PCM+SR-LUT & 28.04\textcolor{red}{(+1.33)} & 28.55\textcolor{red}{(+1.39)} & 27.33\textcolor{red}{(+1.29)} & 28.78\textcolor{red}{(+0.77)} & 32.33\textcolor{red}{(+2.96)}\\
        MuLUT\cite{li2022mulut} & 28.11 & 29.01 & 27.67 & 29.88 & 33.24\\
        PCM+MuLUT & 29.17\textcolor{red}{(+1.05)} & 29.92\textcolor{red}{(+0.91)} & 28.74\textcolor{red}{(+1.07)} & 30.33\textcolor{red}{(+0.44)}& 34.33\textcolor{red}{(+1.09)}\\
        RC-LUT\cite{liu2023reconstructed} & 28.12 & 29.07 & 27.80 & 29.89 &33.88\\
        PCM+RC-LUT & 29.22\textcolor{red}{(+1.10)} & 30.09\textcolor{red}{(+1.02)} & 28.77\textcolor{red}{(+0.97)} & 30.54\textcolor{red}{(+0.65)}& 34.67\textcolor{red}{(+0.79)}\\
        SPF-LUT\cite{li2024look} & 28.56 & 29.58 & 28.26 & 30.44 & 34.91\\
        PCM+SPF-LUT & 29.72\textcolor{red}{(+1.16)} & 30.63\textcolor{red}{(+1.05)} & 29.43\textcolor{red}{(+1.17)}& 30.96\textcolor{red}{(+0.52)} & 35.56\textcolor{red}{(+0.65)}\\
        \bottomrule[1.1pt]
        
    \end{tabular}
    \vspace{-1.4em}
    \label{tab:pcm}
\end{table*}

\subsection{Implementation details}
We implement DnLUT using the Adam optimizer with a cosine annealing learning rate schedule and mean squared error (MSE) loss function. For Gaussian denoising, we train networks for $2 \times 10^5$ iterations with a batch size of 12, increasing to 32 for real-world denoising scenarios. Following established practices\cite{jo2021practical, li2022mulut}, we uniformly sample LUTs at $2^4$ intervals. To minimize indexing artifacts, final predictions utilize 4D simplex interpolation and 3D tetrahedral interpolation. We further refine cached LUTs through 2000 iterations of LUT-aware fine-tuning\cite{li2022mulut} on the training dataset.

\subsection{Gaussian Color Image denoising}
Following \cite{zamir2022restormer}, we train our model using four comprehensive datasets: DIV2K\cite{Agustsson_2017_CVPR_Workshops}, Flickr2K\cite{timofte2017ntire}, BSD400\cite{MartinFTM01}, and WaterlooED\cite{ma2016waterloo}. Training data is augmented with additive Gaussian noise at three levels ($\sigma \in \{15, 25, 50\}$). We evaluate performance on four standard benchmarks: CBSD68\cite{martin2001database}, Kodak24, Urban100\cite{huang2015single}, and McMaster\cite{zhang2011color}, using identical noise levels.

\subsection{Real-world Color Image denoising}
To validate real-world performance, we train DnLUT on the SIDD training dataset and evaluate on both SIDD validation and DnD datasets. The SIDD dataset comprises indoor scenes captured by five smartphones under varying noise conditions, providing 320 image pairs for training and 1,280 for validation. The DnD dataset offers 50 high-resolution images with realistic noise patterns for comprehensive evaluation.

\subsection{Quantitative results}
We evaluate performance using color peak signal-to-noise ratio(CPSNR) and structural similarity index(SSIM) metrics, comparing against state-of-the-art methods across three categories: LUT-based methods (SR-LUT\cite{jo2021practical}, MuLUT\cite{li2022mulut}, RC-LUT\cite{liu2023reconstructed}, SPF-LUT\cite{li2024look}), classical approaches (CBM3D\cite{dabov2007color}, MC-WNNM\cite{xu2017multi}), and DNN-based solutions (DnCNN\cite{zhang2017beyond}, SwinIR\cite{liang2021swinir}). 

As shown in Tab.~\ref{tab:benchmark}, DnLUT achieves superior performance in the LUT-based category, surpassing SPF-LUT by over 1dB in Gaussian noise removal while requiring only 17\% of its storage. Results from Tab.~\ref{tab:sidd} demonstrate DnLUT's effectiveness in real-world scenarios, outperforming classical methods CBM3D\cite{dabov2007color} and MC-WNNM\cite{xu2017multi} by approximately 5dB. Though the Gaussian noise is independent across RGB channels, the network can still leverage inter-channel natural distribution in images to suppress the noise. For example, if a pixel in the R channel has a high value due to noise, but the corresponding pixels in the G and B channels do not, the network can infer that the high value is likely noise and suppress it.

\subsection{Qualitative results}
Visual comparisons in Figs.~\ref{fig:Gua} \&~\ref{fig:sidd} illustrate DnLUT's superior denoising capabilities. Existing LUT-based methods exhibit noticeable color distortion, particularly evident in Fig.~\ref{fig:Gua}'s first row. In contrast, DnLUT achieves visual quality comparable to computation-intensive methods like DnCNN and CBM3D. Fig.~\ref{fig:sidd} further demonstrates DnLUT's advantage in real-world scenarios, where both LUT-based and classical methods struggle with color fidelity and often over-smooth details. DnLUT successfully preserves color consistency while retaining fine image details.

\begin{table*}[htp]
    \centering
    \caption{Efficiency evaluation of three categories color image denoising methods. We use the Qualcomm Snapdragon 8 Gen 2 as the mobile platform. The energy cost is calculated on the $512 \times 512$ resolution color images.}
    \begin{tabular}
    {p{15mm}<{\centering}  p{28mm}<{\centering} p{20mm}<{\centering}  p{25mm}<{\centering}  p{25mm}<{\centering}  p{15mm}<{\centering}  p{20mm}<{\centering}}
    \toprule[1.1pt]
      \multirow{2}{*}{Category}&\multirow{2}{*}{Method} &\multirow{2}{*}{Platform}  &Runtime(ms)& Runtime(ms) &Energy  & Storage  \\
      &&& $256\times256$& $512\times512$&Cost (pJ)&(KB) \\
      \hline
       \multirow{5}{*}{LUT-based} &SR-LUT\cite{jo2021practical} &Mobile &  6 & 19 &149.98M & 82\\
        &MuLUT\cite{li2022mulut} & Mobile &  21 & 80 & 899.88M & 489\\
        &RC-LUT\cite{gharbi2017deep} & Mobile &  18 & 78 & 612.92M & 326\\
        &SPF-LUT\cite{li2024look} & Mobile& 72 & 257 & 2.32G &30,178 \\
        &DnLUT(ours) & Mobile & 23  & 88 & 687.34M & 518\\
        \midrule
        \multirow{2}{*}{Classical}&CBM3D\cite{zhang2022clut} & PC & 3,189 & 14,343& 4.82G & -\\
        &MC-WNNM\cite{xu2017multi}  & PC & 74,290 & 293,424  & 89.23G & -\\
        \midrule
        \multirow{2}{*}{DNN}&DnCNN\cite{zhang2017beyond} & Mobile & 543 & 1,923 &542.53G & 2,239\\
        &SwinIR\cite{liang2021swinir} & Mobile & 73,234 & 274,453&12.03T & 45,499\\
      \bottomrule[1.1pt]
      
    \end{tabular}

    \label{tab:burden}
    \vspace{-1em}
\end{table*}

\subsection{Efficiency Evaluation}
We conduct comprehensive efficiency analysis across three key metrics: theoretical energy cost, runtime performance, and storage requirements (Tab.~\ref{tab:burden}).

\noindent \textbf{Energy cost.} Following AdderSR\cite{song2021addersr}, we analyze multiplication and addition operations across different data types to estimate total energy consumption. DnLUT achieves 70\% energy savings compared to SPF-LUT while delivering superior denoising quality. The efficiency gap widens dramatically when compared to DNN-based methods, with DnLUT requiring only 0.1\% of DnCNN's energy consumption.

\noindent \textbf{Runtime.} We implement LUT-based methods using standard JAVA API on the ANDROID platform. DnLUT maintains the rapid inference capabilities characteristic of SR-LUT\cite{jo2021practical}, demonstrating clear advantages over classical and DNN-based approaches. For reference, DnCNN running on mobile CPU requires 20× more processing time. DnLUT's performance can be further optimized through device-specific implementations, such as FPGA deployments.

\noindent \textbf{Storage.} DnLUT requires approximately 500 KB storage, well within typical L2-cache limits of modern chips. This efficient memory footprint enables rapid lookup operations with minimal dependency overhead.

\subsection{Effectiveness of Plug-in PCM}
\begin{figure}
    \centering
    \includegraphics[width=0.98\linewidth]{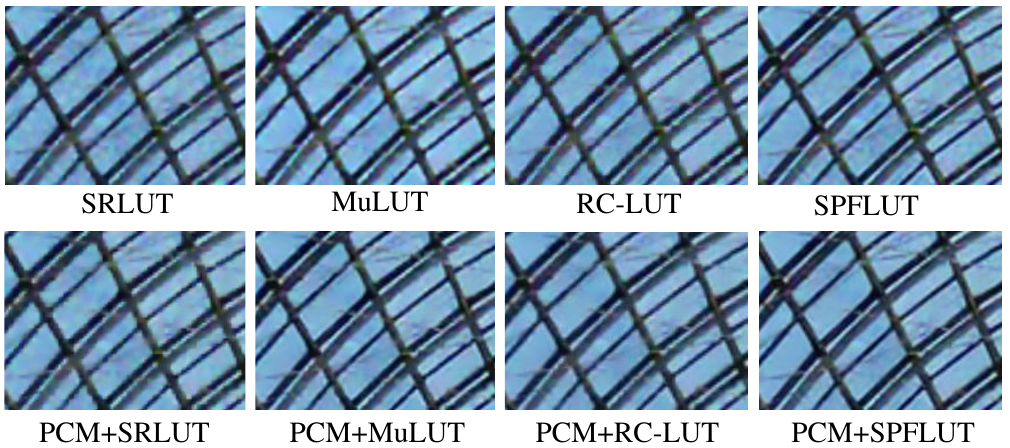}
    \caption{Visualization of LUT-based method with plug-in PCM.}
    \label{fig:PCM}
    \vspace{-1.4em}
\end{figure}

Tab.~\ref{tab:pcm} demonstrates PCM's versatility as a performance enhancer for existing LUT-based methods. When integrated with current approaches, PCM consistently delivers significant improvements across all benchmark datasets. The visual comparisons in Fig.~\ref{fig:PCM} show marked reduction in color distortion, particularly in smooth image regions, resulting in superior perceptual quality.

\subsection{Ablation study}
We conduct ablation studies using Gaussian color image denoising benchmark to validate DnLUT's key components.

\subsubsection{Channel-spatial-wise convolution}
Channel indexing\cite{li2024toward,yang2024taming} possesses a $1\times1$ receptive field to extract channel information. We evaluate various combinations of channel-wise, spatial-wise, and our proposed channel-spatial-wise kernels using a consistent architecture comprising one channel indexing module, two L-shaped convolution modules, and one PCM component.

Results in Tab.~\ref{tab:indexing} reveal that naive combination of channel-wise and spatial-wise convolutions yields suboptimal performance. Pure channel-spatial-wise convolution, while promising, struggles with limited receptive field coverage. The integration of spatial-wise and channel-spatial-wise convolutions produces substantial improvements. Notably, adding channel-wise convolution offers minimal additional benefit, as channel-spatial-wise convolution already captures comprehensive color information.

\vspace{-0.2em}
\subsubsection{L-shaped convolution}
To validate our L-shaped convolution design, we compare it against standard $2\times2$ spatial-wise kernels (denoted as S in Tab.~\ref{tab:l}, $\sigma = 1$). While the conventional kernel shows marginal performance advantages, our L-shaped design achieves comparable results while enabling 3D LUT conversion, reducing storage requirements by 17×. It paves the way of deploying multi-layer spatial-wise convolutions. 

\begin{table}[htp]
    \centering
    \caption{Different combinations of channel-wise, spatial-wise, and our channel-spatial-wise kernels. These kernels are represented by channel indexing\cite{li2024toward}, L-shaped kernel, and PCM. The table presents CPSNR(dB) values.}
    \vspace{-0.8em}
    \begin{tabular}{p{15mm}<{\centering}  p{15mm}<{\centering}  p{25mm}<{\centering}  p{12mm}<{\centering}  }
        \toprule[1.1pt]
        Channel & Spatial & Channel-spatial &CBSD68 \\
        \hline
        & \checkmark & & 29.77\\
        & & \checkmark & 30.15\\
        \checkmark & \checkmark & & 30.21\\
        & \checkmark & \checkmark & 31.41\\
        \checkmark & \checkmark& \checkmark & 31.47\\
        \bottomrule[1.1pt]
        
    \end{tabular}
    \vspace{-0.9em}
    \label{tab:indexing}
\end{table}

\begin{table}[htp]
    \centering
    \caption{Comparison of L-shaped kernel and $2\times2$ spatial-wise kernel. The table presents CPSNR(dB) values.}
    \vspace{-0.8em}
    \resizebox{1\columnwidth}{!}{
    \begin{tabular}{ccccc}
        \toprule[1.1pt]
       Method& CBSD68 & Kodak&McMaster & LUT Size(KB) \\
            \hline
        PCM~+~L  & 26.63 & 27.05 & 27.96 &494\textcolor{red}{~+~24}\\
        PCM~+~S  & 26.71 & 27.16 & 28.01 &494\textcolor{red}{~+~408}\\
      \bottomrule[1.1pt]
      
    \end{tabular}}
    \vspace{-0.6em}
    \label{tab:l}
\end{table}


\section{Conclusion}

This paper presents DnLUT, a novel approach to color image denoising that bridges the gap between high-performance deep learning models and resource-constrained edge devices. Addressing the limitations of existing LUT-based methods, DnLUT introduces two key innovations. First, the Pairwise Channel Mixer (PCM) models channel and spatial correlations simultaneously by processing channel pairs, enabling comprehensive color modeling without excessive storage costs. Second, an L-shaped convolution design rethinks rotation-based receptive field expansion, eliminating redundant pixel usage and converting 4D to 3D LUTs, reducing storage by 17× while maintaining spatial coverage. These advancements allow DnLUT to outperform state-of-the-art LUT-based methods by over 1dB in denoising quality while consuming just 0.1\% of DnCNN's energy.

\vspace{1.6mm}
\noindent\textbf{Acknowledgements}.
This research was supported by: the HKU-TCL Joint Research Centre for AI, the Theme-based Research Scheme (TRS) project T45-701/22-R, and the General Research Fund (GRF) Project 17203224 from the Research Grants Council (RGC) of Hong Kong SAR. Additional support was provided by the National Key Research and Development Program of China (Grant No. 2024YFB2808903), the Shenzhen Science and Technology Program (JCYJ20220818101001004), and the Tsinghua University - Tencent Joint Laboratory for Internet Innovation Technology Research Fund.

\clearpage
\normalem
{
    \small
    \bibliographystyle{ieeenat_fullname}
    \bibliography{main}
}


\end{document}